\documentclass[conference]{IEEEtran}
\usepackage{cite}
\usepackage{amsmath,amssymb,amsfonts}
\usepackage{algorithm}
\usepackage{algpseudocode}
\usepackage{graphicx}
\usepackage{textcomp}
\usepackage{xcolor}
\usepackage{array}
\usepackage{subfigure}
\usepackage{verbatim}
\usepackage{enumitem}
\usepackage{booktabs}
\usepackage{multirow}

\newcommand{\deep}{SaneDL} 
\makeatletter

\newcommand{\Rmnum}[1]{\expandafter\@slowromancap\romannumeral #1@}
\makeatother

\def\BibTeX{{\rm B\kern-.05em{\sc i\kern-.025em b}\kern-.08em
    T\kern-.1667em\lower.7ex\hbox{E}\kern-.125emX}}
\begin{document}

\title{Data Sanity Check for Deep Learning Systems via Learnt Assertions}
\author{
\IEEEauthorblockN{Haochuan Lu\IEEEauthorrefmark{1}{$^\dag$}, Huanlin Xu\IEEEauthorrefmark{1}{$^\dag$}, Nana Liu\IEEEauthorrefmark{1}{$^\dag$}, Yangfan Zhou\IEEEauthorrefmark{1}{$^\dag$}, Xin Wang\IEEEauthorrefmark{1}{$^\dag$}}

\IEEEauthorblockA{\IEEEauthorrefmark{1}School of Computer Science, Fudan University, Shanghai, China} 
\IEEEauthorblockA{\IEEEauthorrefmark{2}Shanghai Key Laboratory of Intelligent Information Processing,
Shanghai, China} 
}
\maketitle
 
\begin{abstract}
Reliability is a critical consideration to DL-
based systems. But the statistical nature of DL makes it
quite vulnerable to invalid inputs, i.e., those cases that are not
considered in the training phase of a DL model. This paper
proposes to perform data sanity check to identify invalid inputs,
so as to enhance the reliability of DL-based systems. We design and implement a tool to
detect behavior deviation of a DL model when processing
an input case. This tool extracts the data flow footprints and conducts
an assertion-based validation mechanism. The assertions are
built automatically, which are specifically-tailored for DL model
data flow analysis. Our experiments conducted with real-world
scenarios demonstrate that such an assertion-based data sanity
check mechanism is effective in identifying invalid input cases.
\end{abstract} 

\section{Introduction}
In recent years, deep learning (DL) techniques have shown
great effectiveness in various aspects. A huge amount of
DL-based applications and systems have been proposed in
favor of people’s daily life and industrial production\cite{DBLP:conf/iccais/DuanM15, DBLP:journals/tmi/GreenspanGS16, DBLP:conf/kbse/AbdessalemNBS16},
even in safety-critical applications. 
Reliability is hence of great significance for practical DL-
based systems.

It is widely-accepted that every software system has its valid
input domain\cite{white1980domain, DBLP:conf/kbse/TanejaLMXT10, li2010perturbation, DBLP:conf/issta/TsankovDB13 }. Inputs staying in such a domain, namely,
valid inputs, are expected to receive proper execution results.
Unfortunately, in real circumstances, there is no guarantee the
inputs are always valid. Anomalous, unexpected inputs may
arrive and result in unpredictable misbehavior, which in turn
degrades reliability. In particular, this is a severe threat to
the reliability of DL-based systems. The statistical-inference
nature of DL models makes them quite vulnerable to invalid
input cases.


In this paper, we propose \textbf{SaneDL} , a tool that provides systematic
data \textbf{San}ity ch\textbf{e} ck for \textbf{D}eep \textbf{L}earning-based Systems.
SaneDL serves as a lightweight, flexible, and adaptive plugin
for DL models, which can achieve effective detection of invalid
inputs during system runtime. The key notion of SaneDL design
is that a DL-model will behave differently given valid input
cases and the invalid ones. The behavior deviation is the
symptom of invalid input cases \cite{engler2001bugs}. Since a DL model is not
complicated in its control flow, we use data flow to model
its behaviors. \deep\ provides an assertion-based mechanism.
Such assertions, typically, are constructed via AutoEncoder
\cite{hinton2006reducing}, which exhibits a perfect compatibility with the DL
model. Similarly to traditional assertions for general programs
inserted between code blocks, the assertions are inserted in
certain network layers, so as to detect behavior deviations in
a data flow perspective. Invalid input cases are thus identified
effectively.

We summarize the contributions of this paper as follows.
\begin{itemize}
\item We approach reliability enhancement of DL systems via
data sanity check. We proposed a tool, namely SaneDL, to perform data sanity check for DL-based systems. SaneDL provides assertion mechanism to detect behavior deviation of DL model.
To our knowledge, SaneDL is the first assertion-based tool
that can automatically detects invalid input cases for DL
systems. Our work can shed light to other practices in
improving DL reliability.

\item SaneDL adopts a non-intrusive design that will not degrade
the performance of the target DL model. In other words,
the SaneDL mechanism does not require the adaptation of
the target DL model. As a result, SaneDL can be easily
applied to existing DL-based systems.
\item We show the effectiveness and efficiency of invalid input
detection via real-world invalid input cases, other than
manipulated faulty samples typically used in the com-
munity. This proves the SaneDL mechanism is promising.
\end{itemize}

We will then discuss our approach in Section II and the evaluation in Section III. 
Some related work are listed in Section IV. We conclude this work in Section V. 
\section{Methodologies}

\begin{figure}[htbp]
	\centering
	\includegraphics[width=1\linewidth]{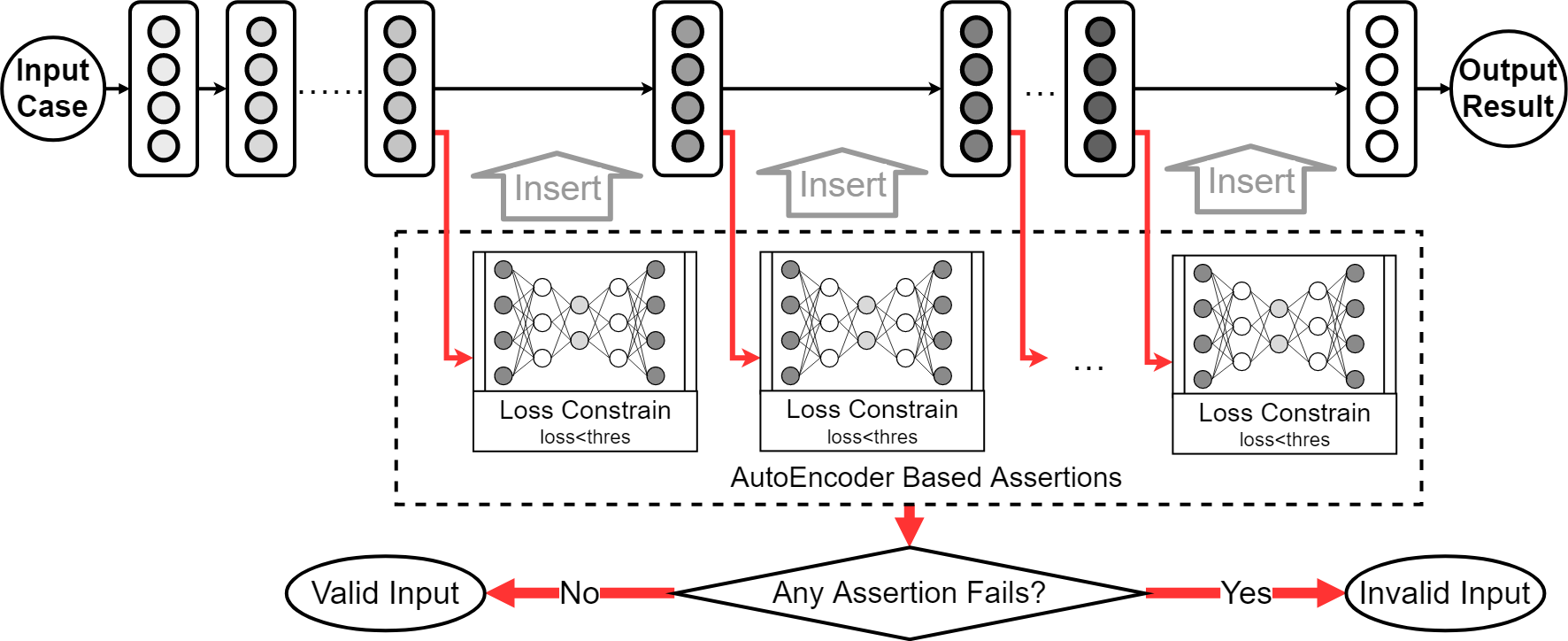}  
	\caption{Overview of \deep} 
	\label{fig:overview}  
\end{figure}

Figure \ref{fig:overview} shows the framework of \deep. Generally, for a pre-trained DNN model, the execution process of \deep\ follows the following workflow.

First, a series of AutoEncoder (AE)\cite{hinton2006reducing} based assertions are inserted into the structure of a pre-trained DNN. Specifically, given a pre-trained neural network $N$, we insert AE networks between its layers. 
These AEs are trained using the intermediate result generated by each layer of $N$. 

Then,  for a given input, such assertions take the intermediate results of neural network as inputs to evaluate the anomalous degree via the losses. AEs can reconstruct all intermediate results from valid inputs. For valid inputs, the intermediate results can be properly restored by AEs. But for invalid inputs, the intermediate results of invalid puts show abnormal patterns. Hence the reconstruction leads to a huge deviation with a high loss value that don’t fit the AEs.


We check each assertions by using properly preset thresholds to constrain the losses. Formally, it can be expressed as the follow equation:
$$
Thres = \delta \cdot \frac{\sum_{j=1}^{m} Loss_{AE}(IR_{j})}{m}
$$
The $IR_{j}$ stands for the input of the AE, which is the intermediate result at certain layer of the $j$th case in the training set. The $m$ means the total samples of the training set. The $\delta$  represents the scale coefficient, which determines the assertion threshold as a certain scale of the mean loss of valid inputs. Our practice shows that selecting this scale coefficient in a interval from 2 to 4 ({\em i.e.} $\delta \in \left [ 2, 4 \right ]$) results in good performance.

Lastly, the assertion results then determine the validity of given input. Only those inputs that pass all the assertions({\em i.e.} the losses don't exceed any constrains) are verified as valid inputs that the deep learning system can properly handle.

\section{Experiment}


\subsection{Experimental Setup}

We build two experimental target scenarios to evluate the effectiveness of \deep.
For Scenario \Rmnum{1}, we build a test set with a combination of cases from Fashion-MNIST and MNIST\footnote{Fashion-MNIST is a public dataset providing pictures of fashionable clothing. MNIST is another public dataset consists of handwritten digits images.}, so as to test if \deep\ can successfully enhance a pre-trained model on Fashion-MNIST, {\em i.e.} to detect invalid inputs from MNIST dataset. Similarly, for Scenario \Rmnum{2}, we train traffic sign recognition models based on GTSRB\footnote{GTSRB stands for German Traffic Sign Recognition Benchmark, which is a dataset consists of traffic signs of different categories}\cite{Houben-IJCNN-2013}, and then test the embedded \deep\ with a hybrid test set. Such hybrid test set is built by combining extra cases from FlickrLogos-27\footnote{FlickrLogos-27 is dataset containing real images of various brand logos that appear in the wild}\cite{DBLP:conf/mir/KalantidisPTZA11} with the original test. We test if \deep\ can figure out those invalid inputs of commercial signs from traffic signs.  

We build pre-trained models in different network structures. For Fashion-MNIST, we use LeNet\cite{lecun1998gradient} and AlexNet\cite{krizhevsky2012imagenet}. And for GTSRB, the AlexNet and VGG16\cite{corr/simonyan2014very} are utilized. We consider the True Positive Rate(TPR) and the False Positive Rate(FPR) as two proper metrics to evaluate the performance of \deep\ in detecting invalid inputs. Besides, the ROC-AUC score is also considered as one of our evaluation metrics to evaluate the overall performance under different $\delta$.

Table \ref{Fashion-Mnist} shows the detection results in two scenarios. For each test setting, we inject invalid cases by randomly replacing some cases in the original test set with the invalid input cases. The results exhibits remarkable performance of \deep. 


\begin{table}
	\caption{Evaluation results on invalid input detection in Scenario \Rmnum{1}}
	\centering
	
	\begin{tabular}{|c|c|c|c|c|c|}
		\hline
		\multicolumn{2}{|c|}{\textbf{Dataset}} & \multicolumn{2}{c|}{\textbf{Fashion-Mnist \& Mnist}} & \multicolumn{2}{c|}{\textbf{GTSRB \& Flickr}}\\
		\hline
		\multicolumn{2}{|c|}{\textbf{Model}} & \textbf{LeNet} & \textbf{AlexNet} & \textbf{AlexNet} & \textbf{VGG} \\
		\hline
		\multicolumn{2}{|c|}{\textbf{Invalid Inputs}} & \multicolumn{2}{c|}{\textbf{Hand-write Digits}} & \multicolumn{2}{c|}{\textbf{Commercial Signs}} \\
		\hline
		\multirow{3}{*}{\textbf{Metrics}} & \textbf{TPR} & 0.9919 & 0.9975 & 0.9451 & 0.9268\\
		~ & \textbf{FPR} & 0.0156 & 0.0221 & 0.1002 & 0.0801\\
		~ & \textbf{AUC} & 0.9978 & 0.9987 & 0.9808 & 0.9716\\
		\hline
	\end{tabular}
	\label{Fashion-Mnist}
\end{table}
\section{Related Work}

\subsection{Input Validation and Sanitization in Software}

In real production environment, input validation is usually required to prevent unexpected inputs from violating system's pre-defined functionalities. Existing approaches seek for ways that better perform sanity check for specific scenarios\cite{hayes1999increased, DBLP:conf/icse/Avancini12, DBLP:conf/issta/AlkhalafAB14, DBLP:conf/edo/BuehrerWS05, DBLP:conf/sp/BalzarottiCFJKKV08, DBLP:conf/kbse/SharT12, DBLP:conf/compsac/ScholteRBK12, park2008web }. 
All these work shows input validation plays an important role in maintaining software reliability.
However, few experiences on applying input validation to deep learning systems. \deep\ is hence designed to achieve this functionality, so as to enhance the reliability of deep learning systems.

\subsection{Deep Learning Testing}

Recent studies in SE community realize the inadequacy of traditional deep learning testing via statistical results({\em e.g.} accuracy), which can hardly expose the specific defects of neural network. Therefore, several work tends to regard deep neural network as a sophisticated software and apply systematic testing mechanisms according to software engineering domain knowledge\cite{DBLP:journals/corr/abs-1803-04792, DBLP:journals/corr/abs-1807-10875, DBLP:conf/sigsoft/MaLLZG18, DBLP:journals/corr/abs-1710-00486, DBLP:conf/issre/MaZSXLJXLLZW18, DBLP:conf/kbse/MaJZSXLCSLLZW18, DBLP:conf/kbse/SunWRHKK18}. On the other hand, test case generation turns out to be another effective approach to extend the testing effect in deep learning systems to enhance performance and reliability\cite{DBLP:conf/kbse/ZhangZZ0K18, DBLP:conf/sosp/PeiCYJ17, DBLP:conf/icse/TianPJR18}. However, excessive manipulations can violate the internal feature of the original cases. This work more considers deep learning systems' deployment in real practice, where true real-world cases can be encountered instead of manipulated ones.

\section{Conclusions}

Deep learning systems exhibit huge effectiveness in real life. But, it may suffer from invalid inputs encountered in complex circumstances. 
This work presents our efforts on handling invalid inputs that deep learning systems may come across in practice. We propose a white-box verification framework, namely, \deep\, to perform systematic data sanity check for deep learning systems via assertion-based mechanism. 
It works in an efficient, non-intrusive manner to detect invalid inputs. 
Our experiment shows its good performance when being applied to two practical problem scenarios, where the invalid input cases are real-world cases rather than manually-generated toy cases. 


 
\newpage

\bibliographystyle{IEEEtran}
\bibliography{aitest}    

\end{document}